\documentclass{article} % For LaTeX2e
\usepackage[preprint,nonatbib]{nips_2018}
\usepackage[numbers]{natbib}
\usepackage{times}
\usepackage{amsmath,amssymb}
\usepackage{hyperref}
\usepackage{graphicx}
\usepackage{subcaption}
\usepackage{url}
\usepackage{color}
\usepackage{ulem}

\definecolor{orange}{rgb}{1,0.5,0}
\definecolor{darkgreen}{rgb}{0,0.5,0}
\definecolor{red}{rgb}{1,0,0}

\title{Generating Diverse Programs with Instruction Conditioned Reinforced Adversarial Learning}

\author{
  Aishwarya Agrawal \\
  Georgia Tech\thanks{This work was done while Aishwarya was interning at DeepMind.}\\
  \texttt{aishwarya@gatech.edu} \\
  \And
  Mateusz Malinowski \\
  DeepMind\\
  \texttt{mateuszm@google.com} \\
  \And
  Felix Hill \\
  DeepMind\\
  \texttt{felixhill@google.com} \\
  \And
  Ali Eslami \\
  DeepMind\\
  \texttt{aeslami@google.com} \\
  \And
  Oriol Vinyals \\
  DeepMind\\
  \texttt{vinyals@google.com} \\
  \And
  Tejas Kulkarni \\
  DeepMind\\
  \texttt{tkulkarni@google.com} \\
}

\begin{document}

\maketitle

\begin{abstract}
Advances in Deep Reinforcement Learning have led to agents that perform well across a variety of sensory-motor domains. In this work, we study the setting in which an agent must learn to generate programs for diverse scenes conditioned on a given symbolic instruction.
Final goals are specified to our agent via images of the scenes. A symbolic instruction consistent with the goal images is used as the conditioning input for our policies. 
Since a single instruction corresponds to a diverse set of different but still consistent end-goal images, the agent needs to learn to generate a distribution over programs given an instruction. 
We demonstrate that with simple changes to the reinforced adversarial learning \cite{ganin2018synthesizing} objective, we can learn instruction conditioned policies to achieve the corresponding diverse set of goals. Most importantly, our agent's stochastic policy is shown to more accurately capture the diversity in the goal distribution than a fixed pixel-based reward function baseline. 
We demonstrate the efficacy of our approach on two domains: (1) drawing MNIST digits with a paint software conditioned on instructions and (2) constructing scenes in a 3D editor that satisfies a certain instruction. 

\end{abstract}

\section{Introduction}

\begin{figure*}[t]
\begin{center}
\scalebox{0.8}{
\includegraphics[width=\linewidth]{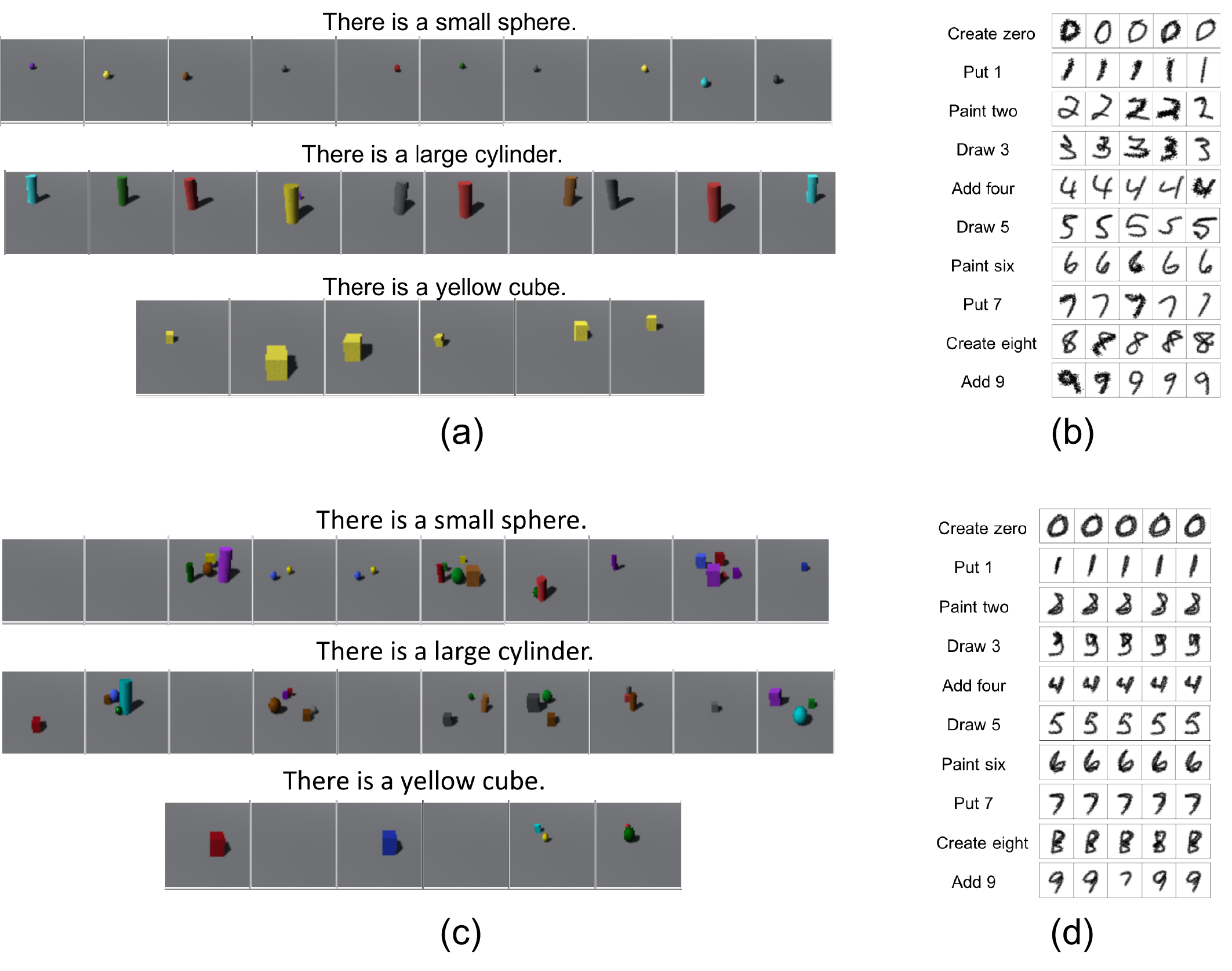}
}
\end{center}
% \vspace{-10pt}
\caption{We build agents that can generate programs for diverse scenes conditioned on a given symbolic instruction. (a) Conditioned on a given instruction (e.g., ``There is a small sphere''.), the agent learns to generate programs for 3D scenes such that all scenes have a small sphere. We can see that the generated scenes show diversity in unspecified attributes (color and location of the sphere). (b) Conditioned on a simple instruction (e.g., ``Draw zero'', ``Paint 1'', etc.), the agent learns to generate programs that draw the corresponding
MNIST digits. We can see that the generated drawings of a given label show diversity in style. (c) and (d): Results corresponding to (a) and (b) respectively for the fixed pixel-based reward function baseline. We can see that the scenes / drawings rendered from the generated programs are either not consistent with the input instruction (c) or do not show sufficient diversity (d).}
\label{fig:teaser}
\end{figure*}

Virtual worlds are being increasingly used to make advancements in deep reinforcement learning research \cite{ganin2018synthesizing, bahdanau2018learning, embodiedqa, DBLP:journals/corr/HermannHGWFSSCJ17}. However, these worlds are typically hand-designed, 
and therefore their construction at scale is expensive.
In this work, we build generative agents that output programs for diverse scenes that can be used to create virtual worlds, conditioned on an  instruction. Thus, with a short text string, we can obtain a distribution over multiple diverse virtual scenes, each consistent with the  instruction. We experiment with two instruction-conditioned domains: (1) drawing MNIST digits with a paint software, and (2) constructing scenes in a 3D editor; both are shown in Figure \ref{fig:teaser}.

% \vspace{-10pt}

We build our agents on top of recently introduced reinforced adversarial learning framework \citep{ganin2018synthesizing}, where final goals are specified to the agent via images of the virtual scenes. Unlike \citep{ganin2018synthesizing} where, either random noise (unconditional generation) or the goal image (reconstruction) is used as the conditioning input, in our work, a symbolic instruction is used as the conditioning input for our policies. Since, a single instruction corresponds to a diverse set of, yet consistent goal images, the agent needs to learn to generate a controlled distribution over programs such that all programs are consistent with the instruction. A discriminator that takes an image (generated or an example goal image) and the corresponding instruction acts as a reward learning function to provide training signal to the generator. Thus, the generator learns to generate the distribution of programs consistent with the symbolic instruction. In particular, this setting leads to more accurate and more diverse programs than hand-crafted discriminators such as L2-distance (in pixel space) between the generated and goal images.

Language conditioned policies with adversarial reward induction has also been explored in \citep{bahdanau2018learning}. However, \citep{bahdanau2018learning} do not evaluate whether the agent is capable of capturing the diversity in the programs. Moreover, we train our agents on more visually complex tasks involving 3D scenes.

Our work also draws inspiration from Visual Question Answering and Analysis-by-Synthesis:

\textbf{Visual Question Answering.} Visual question answering is a well-established subfield of computer vision that tests a machine's understanding of vision, language, spatial arrangement of objects, knowledge based and commonsense reasoning via question-answer pairs about images~\citep{malinowski2014multi,malinowski2014towards,geman2015visual,antol2015vqa,vqa_ijcv,johnson2017clevr,agrawal2017c,agrawal2018don}. Our work draws inspiration from this line of research, especially \cite{johnson2017clevr} by focusing on similar aspects of the scene comprehension. 
% However, our goal is not entirely the same, and the whole problem involves different challenges
However, in our work, the machine's task is not to generate an answer, but to generate a scene by sequentially generating programs. The latter involves very different challenges (e.g., reward learning, modeling diversity in generated outputs, operating in rich action space). 
(e.g., reward learning, modeling diversity in generated outputs, operating in a rich action space).
Finally, we also believe that architectures based on latent programs, or simulators are under-represented in the visual question answering community~\citep{malinowski2014multi,chowdhury2016contextual,johnson2017inferring,DBLP:journals/corr/HuARDS17,mascharkatransparency,wagner2018answering} even though they offer clear benefits such as interpretability, compositionality, and can potentially generalize better to new instances. This aspect of generalization has recently been recognized as one of the major bottlenecks of the current state of affairs~\citep{agrawal2018don}.

\textbf{Analysis-by-Synthesis.}
Representing images or videos by latent programs 
has also shown promise in learning meaningful representations of data~\citep{yi2018neural,nair2008analysis,kulkarni2015picture,kulkarni2015deep,kulkarni2015picture,malinowski2014multi,wu2017learning,ganin2018synthesizing,bahdanau2018learning}. 
Such frameworks not only improve the interpretability of otherwise obscured decision making systems, but also have the potential to generalize better to unseen data, owing to stronger inductive biases. For instance,  \citep{kulkarni2015picture} use latent programs to infer novels poses of 3D objects, \citep{wu2017learning} show how to invert physical simulation which in turn can be used for inferring objects' dynamics, and \citep{yi2018neural} achieve the state-of-the-art on the CLEVR visual question answering problem.  

\paragraph{Contributions.} We (1) extend the recently proposed reinforced adversarial learning framework \cite{ganin2018synthesizing} to instruction conditioned policies, (2) present the experimental results on two instruction-conditioned domains (drawing MNIST digits with a paint software and constructing scenes in a 3D editor), and (3) show that the proposed setting leads to more accurate and more diverse programs than a fixed pixel-based reward function baseline.

\section{Approach}
\label{sec:model}

\begin{figure*}[h]
% \centering
\begin{center}
\scalebox{0.85}{
\includegraphics[width=\linewidth]{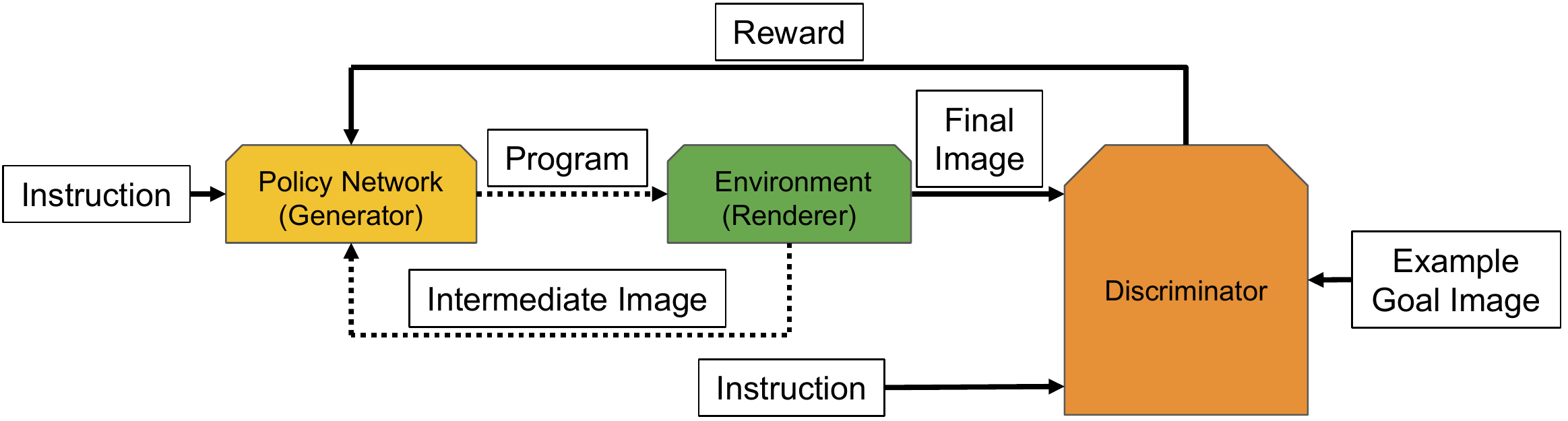}
}
\end{center}
\caption{
An overview of our approach. 
Given an instruction
the generator (policy network) outputs a program which is rendered by a non-differentiable (denoted by dashed arrows) renderer into an image. This process repeats for a fixed number of time steps. The final image is fed to a discriminator, along with the instruction (the same that is given to the generator), that produces a reward.
}
\label{fig:model_pipeline}
\end{figure*}

Our goal is to train a generator agent $G$ that is capable of generating programs for diverse scenes ($x$) conditioned on an instruction $q$ that are indistinguishable from the ground truth data samples $p_d(x|q)$. Towards this goal, we frame our problem within the reinforcement adversarial learning setting~\citep{goodfellow2014generative,mnih2016asynchronous}. 
The generator is modeled by a recurrent policy network  which, at every time step, takes a symbolic instruction as input, and outputs a program which is executed by a renderer to produce an image (Figure \ref{fig:model_pipeline} shows the whole pipeline). For instance, if the instruction is ``Draw zero'', the generator uses the available painting tools (location, pressure, and brush size) to sequentially paint the digit 0 on the canvas. Similarly, with the input ``There is a red sphere'' the generator manipulates the scene by adding / changing objects so that the resulting scene has a red sphere.
The role of the discriminator $D$ is to decide if the distribution over generated images given an instruction ($p_g$) follows the ground truth distribution of images given the instruction ($p_g \approx p_d$).

\begin{figure*}[t]
\begin{center}
\scalebox{0.9}{
\includegraphics[width=\linewidth]{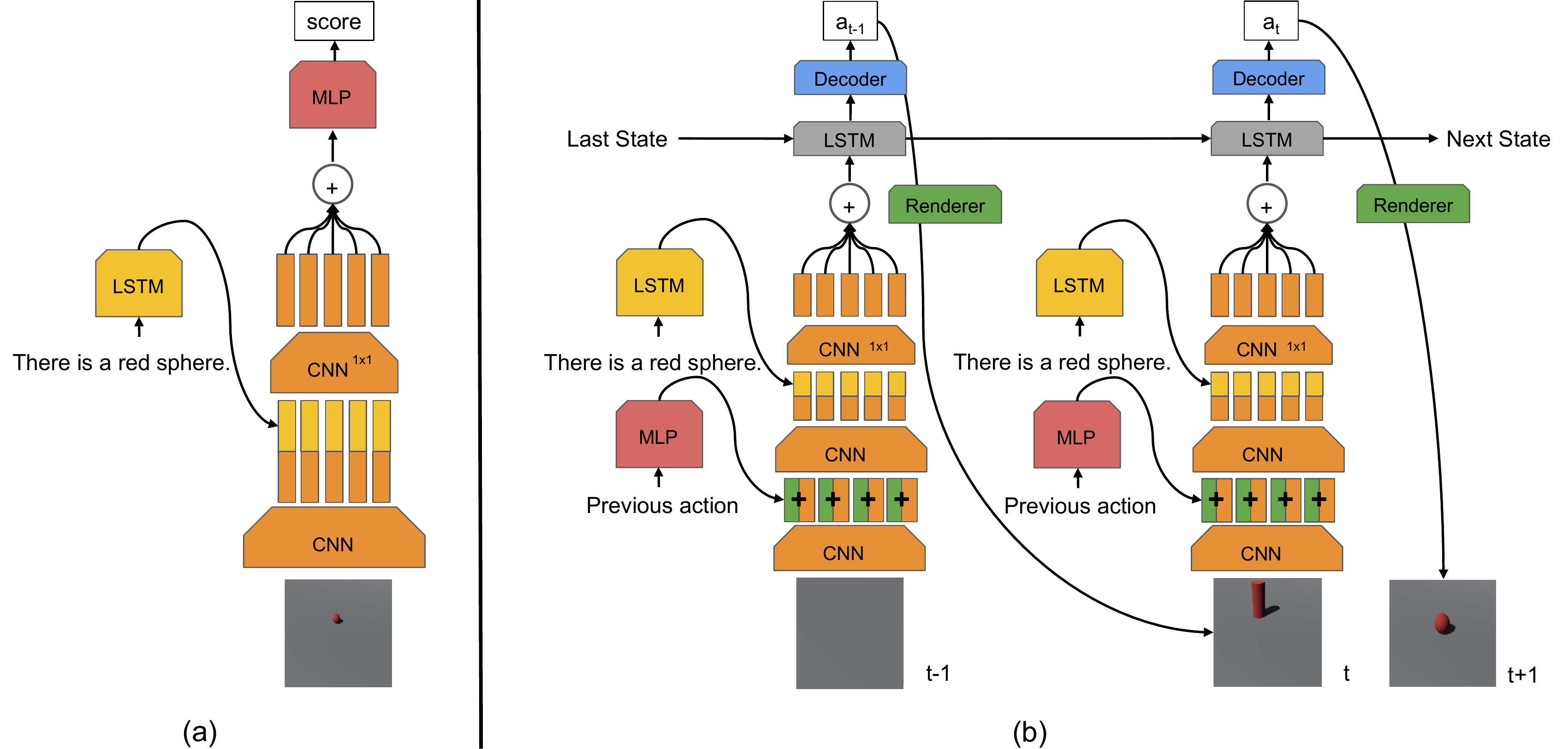}
}
\end{center}
\caption{(a) Discriminator's architecture. It takes as inputs the instruction and an image (either the image generated by the agent or a goal image from the dataset), learns a joint embedding, and outputs a scalar score evaluating the `realness' of the input pair. (b) Policy Network's architecture, unrolled over time. At each time step, it takes -- an instruction, the rendered image from the previous time step (blank canvas initially) and the previous action, learns a joint embedding and passes it to an LSTM whose output is fed to a decoder which samples the actions.}
\label{fig:generator_discriminator}
\vspace{-0.6cm}
\end{figure*}

\textbf{Discriminator.}
To train the discriminator, we use LSTM~\citep{hochreiter1997long} to transform the text instruction into a fixed-length vector representation. 
Using CNNs, we transform the input image into a spatial tensor. 
Inspired by architectures used in visual question answering~\citep{malinowski2018visualqadevil} and text-to-image Generative Adversarial Networks~\citep{DBLP:journals/corr/ReedAYLSL16}, we concatenate each visual spatial vector with the replicated LSTM vector, and process both modalities by 4 layers of 1-by-1 CNNs. 
These representations are then summarized by a permutation-invariant sum-pooling operator followed by an MLP that produces a scalar score which is used as the reward for the generator agent (Figure \ref{fig:generator_discriminator}a).
We train the discriminator by minimizing the minimax objective from~\citep{gan_goodfellow}:
$\mathcal{L}_{D} = -\mathbb{E}_{x \sim p_d} \left[ \log{(D(x|q))} \right] - \mathbb{E}_{x \sim p_g} \left[ \log{(1 - D(x|q))} \right]  + R$, where $R$ denotes gradient penalty, gently enforcing Lipschitz continuity~\citep{gulrajani2017improved}.

\textbf{Generator.} 
Generator's goal is to generate programs that lead to scenes that match data distribution $p_d(x|q)$. Here, programs are defined as a sequence of commands $\mathbf{a} = \left(a_1, a_2, \ldots, a_N\right)$. Every command in this sequence is transformed into scenes through a domain-specific transformation $\mathcal{T}(\cdot)$. For instance, if $\mathcal{T}$ is a 3D renderer, and $\mathbf{a}$ is a program that describes a scene, $\mathcal{T}(\mathbf{a})$ is the image created by executing the program by the renderer. The loss function for the generator is set up adversarially to that for $D$, i.e. $\mathcal{L}_{G} = -\mathbb{E}_{x \sim p_g} \left[ D(x) \right]$.
Due to the non-differentiability of the renderer, we train the network with the REINFORCE~\citep{williams1992simple} and advantage actor-critic (A2C), $\mathcal{L}_{G} = -\sum_t \log \pi(a_t | s_t;\theta) \left[ R_t - V^{\pi}(s_t) \right]$, following~\citep{ganin2018synthesizing}. Figure \ref{fig:generator_discriminator}b depicts the architecture.

\section{Experiments}

\textbf{MNIST digit painting.}
We constructed a dataset consisting of simple, templated instructions such as ``Draw zero.'', ``Paint 5.'' etc. paired with random images for the corresponding MNIST digit.
Each instruction follows the template: $<Action>$ $<Class Label>$ with $Action := Draw \; | \; Put \; | \; Paint \; | \; Add \; | \; Create$ and $<Class Label>$ is specified either numerically (`0') or in word format (`zero'). This dataset has 60K image-instruction pairs (one instruction per image) in total. Following~\citep{ganin2018synthesizing}, we use the `libmypaint' painting library as our rendering environment. The agent produces a sequence of strokes on a canvas $C$ using a brush. Each action involves predicting 5 discrete quantities -- end point and control point of the brush (on 32 x 32 grid), pressure applied to the brush (10 levels), brush size (4 levels) and a binary flag to choose between producing a stroke and skipping (not painting anything). So, the size of the action space is 83,886,080.

Figure~\ref{fig:teaser}b shows samples of MNIST digits painted by our agent when conditioned on the corresponding instructions. We can see that the agent draws the correct digit most of the time and for a given instruction, shows diversity in the generated samples, unlike the fixed pixel-based reward function baseline (L2) whose outputs do not show sufficient diversity (Figure~\ref{fig:teaser}d). Moreover, the baseline takes twice as much time as our agent before it starts generating MNIST-like images.

Quantitatively, to evaluate the correctness (whether the generated samples are consistent with the instruction) and diversity (for a given instruction, whether the generated samples are diverse), we report Inception Score \cite{DBLP:journals/corr/SalimansGZCRC16} (a combined score of quality and diversity) and Fr\'echet Inception Distance (FID) \cite{DBLP:journals/corr/HeuselRUNKH17} (a measures of how close the generated image distribution is to the real image distribution) for 1000 samples using a pretrained MNIST classifier (Table~\ref{tab:mnist_clevr_quant}). We can see that our learned Discriminator outperforms the fixed pixel-based reward baseline (L2).

\begin{table}[h]
\begin{center}
\scalebox{0.90}{
\begin{tabular}{ |c|c|c|c|c|c| } 
\hline
& \multicolumn{2}{c|}{MNIST Digit Painting} & \multicolumn{3}{c|}{3D Scene Construction} \\
Approach & Inception Score & FID & Correctness & Diversity in Colors & Diversity in Sizes \\
\hline
L2 & 1.22 & 283.5 & 1.4 & 0.77 & 0.09 \\ 
Discriminator & \textbf{1.39} & \textbf{259.7} & \textbf{6.8} & \textbf{1.48} & \textbf{0.53} \\ 
\hline
\end{tabular}
}
\end{center}
\caption{
Quantitative evaluation of Discriminator and L2 (in pixel space). The former outperforms the latter on the two domains that we use in our experiments. Depending on the domain, we report Inception Score, Correctness, Diversities -- the higher the better, and FID -- the lower the better.
}
\label{tab:mnist_clevr_quant}
\end{table}

\textbf{3D scene construction.}
Inspired by  CLEVR~\citep{johnson2017clevr}, we constructed a dataset consisting of images of 3D scenes and textual instructions. Each scene consists of a 3D object (cylinder, sphere, cube) of varying sizes, colors that are placed at different locations on the 32 x 32 grid. 
The instructions are generated using templates
of the form ``There is a $<Attribute> <Shape>$'', where attributes are either colors (8 values) or sizes (2 values), and shapes are: cubes, spheres or cylinders. This dataset has 32,318 image-instruction pairs in total. We use a 3D editor as our rendering environment. The agent is provided with an empty scene and it adds objects sequentially. Each action involves predicting 5 discrete quantities -- location of the object (on 32 x 32 grid), object shape, size and color and a flag to choose between adding a new object, changing the previously added object, or doing nothing. So, the size of the action space is 147,456.

Figure~\ref{fig:teaser}(a) shows samples of scenes generated by our agent when conditioned on the corresponding instructions. We can see that the generated scenes are consistent with the instruction and also show diversity in the attribute of the object which is not mentioned in the instruction (e.g., diversity in colors when the instruction is `There is a small sphere.'). The fixed pixel-based reward function baseline (L2) (Figure~\ref{fig:teaser}(c)), fails poorly in following the instruction. 

Quantitatively, we evaluate the correctness and diversity for the unspecified attribute (for samples which are correct) as reported in Table~\ref{tab:mnist_clevr_quant} for 10 samples for each instruction. Each sample is judged as correct / incorrect based on human evaluation. For measuring the diversity, we compute the entropy of the corresponding attribute distribution (number of correct samples for each of the 8 colors / 2 sizes). We can see that our learned Discriminator outperforms the L2 baseline.

\textbf{Acknowledgments:} The authors wish to acknowledge Igor Babuschkin, Caglar Gulcehre, Pushmeet Kohli, Piotr Trochim, and Antonio Garcia for their advice in this project.

\small
\bibliography{biblioLong,biblio}
\bibliographystyle{nips2018_conference}

\end{document}